\documentclass[10pt,twocolumn,letterpaper]{article}

\usepackage[pagenumbers]{cvpr}      

\usepackage{graphicx}
\usepackage{amsmath}
\usepackage{amssymb}
\usepackage{booktabs}
\usepackage{amssymb}
\usepackage{multirow} 
\usepackage{float} 
\usepackage{tikz} 
\usepackage[pagebackref,breaklinks,colorlinks]{hyperref}

\usepackage[capitalize]{cleveref}
\crefname{section}{Sec.}{Secs.}
\Crefname{section}{Section}{Sections}
\Crefname{table}{Table}{Tables}
\crefname{table}{Tab.}{Tabs.}

\def\cvprPaperID{*****} 
\def\confName{CVPR}
\def\confYear{2022}

\begin{document}

\title{Homogeneous and Heterogeneous Consistency Re-ranking for Visible-Infrared Person Re-identification}

\author{Yiming Wang\\
{\tt\small ymwang99@bupt.edu.cn}
}
\maketitle

\begin{abstract}
Visible-infrared person re-identification faces greater challenges than traditional person re-identification due to the significant differences between modalities. In particular, the differences between these modalities make effective matching even more challenging, mainly because existing re-ranking algorithms cannot simultaneously address the intra-modal variations and inter-modal discrepancy in cross-modal person re-identification. To address this problem, we propose a novel Progressive Modal Relationship Re-ranking method consisting of two modules, called heterogeneous and homogeneous consistency re-ranking(HHCR). The first module, heterogeneous consistency re-ranking, explores the relationship between the query and the gallery modalities in the test set. The second module, homogeneous consistency reranking, investigates the intrinsic relationship within each modality between the query and the gallery in the test set. Based on this, we propose a baseline for cross-modal person re-identification, called a consistency re-ranking inference network (CRI). We conducted comprehensive experiments demonstrating that our proposed re-ranking method is generalized, and both the re-ranking and the baseline achieve state-of-the-art performance.
\end{abstract}

\section{Introduction}
\label{sec:intro}

Person re-identification can identify the individuals in pedestrian query sets captured by different cameras by matching unknown identity query sets with known identity gallery sets. Most existing re-identification methods \cite{quan2019autoreid,luo2019bag,huang2019celebrities,cui2023dcr} can only recognize RGB images and perform poorly in low-light or nighttime conditions. To address these issues, some visible-infrared cross-modal person re-identification (V-I ReID) methods \cite{cui2023dcr,wang2020paired,hu2020maximum}  have been proposed, which achieve cross-modal matching by learning the relationships between persons in both visible and infrared modalities.
\begin{figure}[t]
  \centering
   \includegraphics[width=1.0\linewidth]{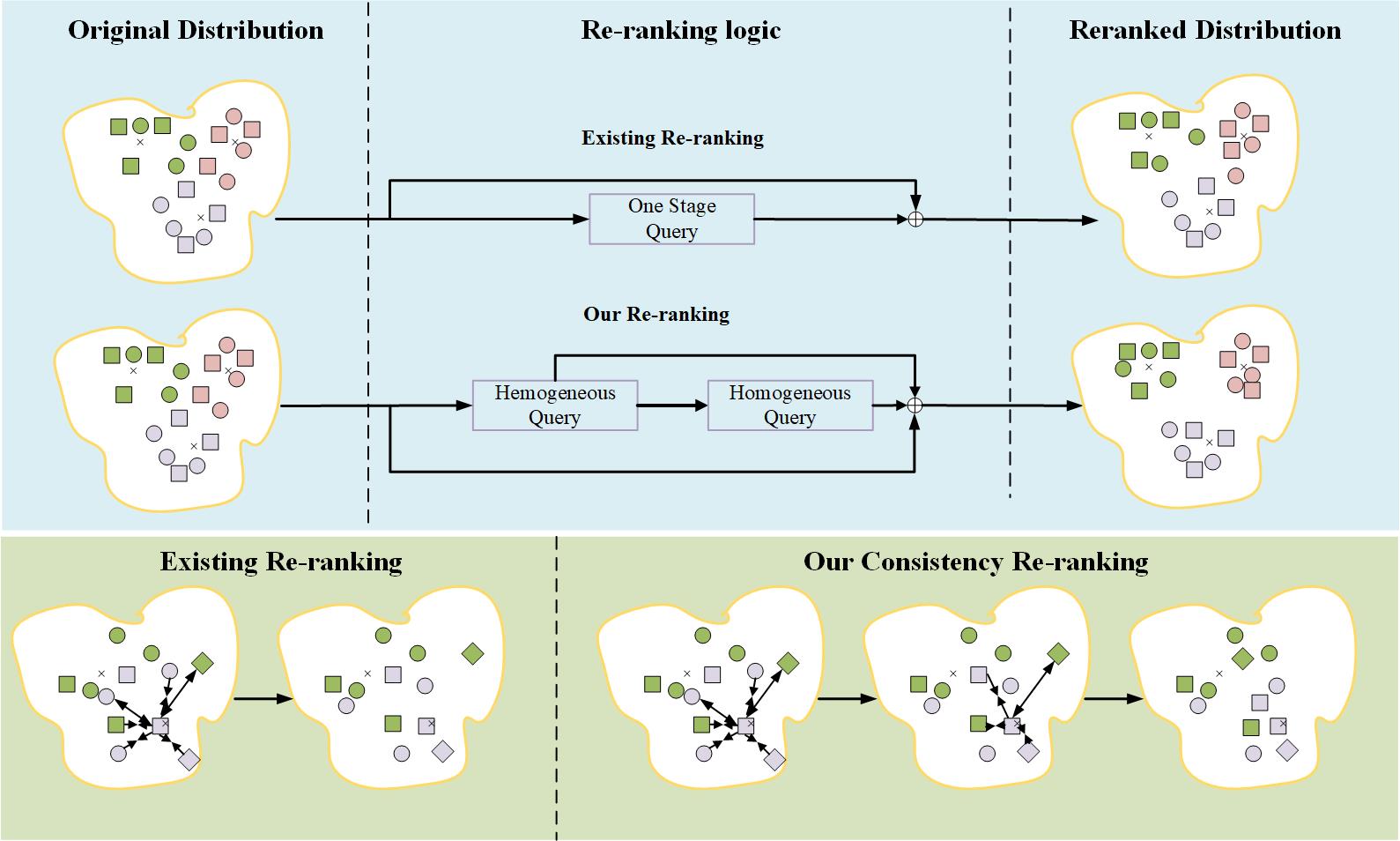}   

   \caption{The crosshair represents the center of the same identity, the same color indicates the same identity, and the circles and rectangles represent the two modalities. The arrows signify the process of reducing or increasing distances. Currently, the re-ranking methods applied to visible-infrared ReID consist of a single-stage re-ranking process, including only homogeneous re-ranking or heterogeneous re-ranking. The re-ranking method we propose includes Homogeneous Consistency Re-ranking and Heterogeneous Consistency Re-ranking.}
   \label{fig:onecol}
\end{figure}
Visible-infrared person re-identification faces a greater challenge than single-modal person re-identification due to the introduction of an infrared modality and the significant gap between visible and infrared modalities. These methods primarily focus on attention mechanisms, feature embedding, counterfactual intervention, and body shape elimination \cite{vaswani2017attention,wang2019joint,diverse_embedding_expansion_network,li2022counterfactual,feng2023shape}. Most of these approaches concentrate on optimizing network architectures. Additionally, the impact of re-ranking \cite{sarfraz2018pose,zhou2022moving,fang2023semantic} on performance is also significant. Optimizing network structures aims to extract higher-quality features and increase the extraction of effective features. The purpose of re-ranking is to design a feature-matching algorithm that can reassess the similarity between query and gallery images, bringing similar images in the query and gallery closer together.

Images contain a significant amount of noise in nighttime environments due to poor image quality and low light. Traditional re-ranking methods typically employ a single-stage approach, which only analyzes intra-modal or inter-modal differences. Recent methods have driven rapid advancements in traditional person re-identification. However, they cannot simultaneously evaluate subtle changes in pedestrian appearance within and across modalities, leading to potential omissions of fine-grained multimodal details. Consequently, such approaches still exhibit certain limitations in cross-modal person re-identification.

To address the issue of detail loss when processing low-quality images in re-ranking, we propose a two-stage progressive re-ranking method based on Graph Convolutional Networks (GCN), named Homogeneous and Heterogeneous Consistency Re-ranking (HHCR). To ensure that both cross-modal and intra-modal pedestrian information are handled, we designed two modules for HHCR: Homogeneous Consistency Re-ranking and Heterogeneous Consistency Re-ranking.
Since the number of visible and infrared images in the test set is unequal, Heterogeneous Consistency Re-ranking is designed as a pseudo-symmetric retrieval method for handling visible and infrared images. Additionally, to address the alignment of intra-modal homogeneous information, Heterogeneous Consistency Re-ranking can separately process visible and infrared images, extracting the consistency within each modality.

Specifically, our re-ranking approach for person re-identification focuses on matching the cross-modal heterogeneous information consistency and intra-modal homogeneous information consistency, dividing HHCR into two steps. In the first step, we apply the concept of graph networks to separately extract the cross-modal heterogeneous information from the query set and gallery set within the entire test set, reducing the impact caused by unequal numbers of images between the query and gallery sets. In the second step, we extract the homogeneous information within the query and gallery sets to reduce the influence of noise from outlier images. Finally, the homogeneous consistency matrix, heterogeneous consistency matrix, and cosine similarity matrix are weighted to form the final HHCR similarity matrix.

urthermore, based on HHCR, we propose a novel Consistency Re-ranking Inference Network (CRI). CRI is trained to update the parameters with a combination of triplet loss and cross-entropy loss functions. During testing, the features extracted by the backbone are input into HHCR, and based on the re-ranking results, CRI determines the identities of pedestrians in the query set.

The contributions of our work are summarized as follows:

\begin{itemize}
\item We propose a consistency re-ranking inference network (CRI) for VI-ReID to explore the consistency of homogeneous and heterogeneous features.
\item We propose an innovative dual-stage progressive re-ranking method named Homogeneous and heterogeneous consistency re-ranking(HHCR). The method includes two modules, homogeneous consistency re-ranking, and heterogeneous consistency e-ranking, which match pedestrians by considering inter and intra-modality differences.
\item  Extensive experiments demonstrate that our CRI and HHCR achieve state-of-the-art accuracy, and the accuracy of our proposed baseline also reaches state-of-the-art levels.
\end{itemize}
 \begin{figure*}[t]
  \centering
   \includegraphics[width=0.8\linewidth]{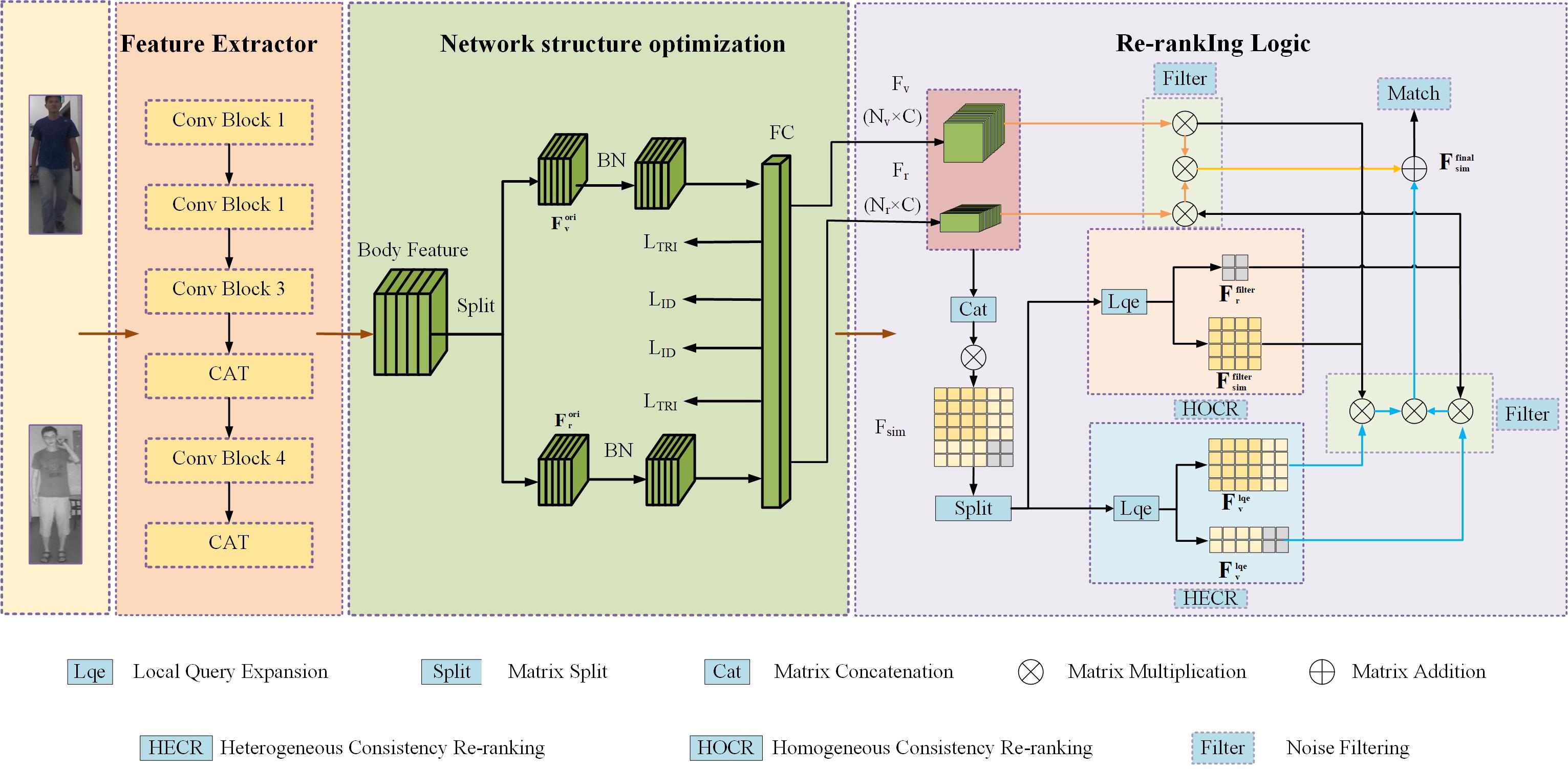}

   \caption{The pipeline of the proposed network. The network structure processes the input visible and infrared images through ResNet to extract features, followed by a BN (Batch Normalization) layer for normalization, and finally, computes the loss function. During the testing phase, HHCR first concatenates the test set to compute the original similarity matrix $A_{sim}^{ori}$. Then, it separately calculates Homogeneous and Heterogeneous Consistency Re-ranking. Finally, the initial similarity matrix is filtered and summed, resulting in the final similarity matrix. The yellow line represents the noise filtering step for the original features, while the blue line indicates the noise filtering step for the original features after HECR graph convolution processing.}
   \label{fig:onecol123}
\end{figure*}
\section{Related Work}
\label{sec:formatting12}

\subsection{Network Structure Optimization For VI-reid}

Cross-modal person re-identification was first proposed to identify pedestrians in nighttime and low-light conditions. The first cross-modal person re-identification network utilized a CNN architecture based on a zero-padding method\cite{wu2017rgb}.

Cross-modal person re-identification was first proposed to identify pedestrians in nighttime and low-light conditions. The first cross-modal person re-identification network utilized a CNN architecture based on a zero-padding method\cite{wu2017rgb}.

Following its introduction, many researchers conducted studies on cross-modal person re-identification, implementing various improvements to the network structure. The first cross-modal person re-identification network based on GANs\cite{goodfellow2014gan}, known as the Cross-Modality Generative Adversarial Network (cmGAN)\cite{dai2018cross}, was designed. This network iteratively optimizes parameters through a minimax game between a generator and a discriminator, where auxiliary information helps bridge the gaps between modalities and facilitates the learning of associations across different modalities. A bi-directional training strategy (BDTR)\cite{ye2018visible} was proposed, allowing the network to learn inter-modal features directly from the data and simultaneously learn both cross-modal and intra-modal relationships to enhance feature representation. A method employing two alignment strategies—pixel alignment and feature alignment—was introduced, marking the first approach to reduce inter-modal and intra-modal variations by modeling these two alignment strategies\cite{wang2019joint}.

Additionally, a Dynamic Dual-Attentive Aggregation (DDAG)\cite{ye2020dynamic} Learning method was proposed, introducing two types of graph attention representations: one for cross-modal graph attention structures that enhance feature representation and another for inter-modal attention that adaptively allocates weights to different body parts. A method for cross-modality shared specific feature transfer (cm-SSFT)\cite{lu2020cross} algorithm has been proposed, effectively leveraging each sample's shared and specific information to improve matching accuracy in cross-modal re-identification tasks. A baseline for cross-modal person re-identification (AGW)\cite{ye2021deep} was introduced, which inserts Non-local Attention\cite{wang2018nonlocal} into the ResNet\cite{he2016deep} architecture. Additionally, a novel re-identification evaluation metric, mINP (mean Inverse Negative Penalty), was proposed to measure the model's ability to capture the most challenging sample features.

An orthogonal decomposition method was developed to separate human features into shape-related and shape-irrelevant features, jointly optimizing shape-related and shape-erasure objectives through shape-guided methods\cite{feng2023shape}. DEEN\cite{diverse_embedding_expansion_network} introduced a novel Diversified Embedding Expansion Module, which facilitates embedding different feature representations, effectively addressing the challenges associated with smaller dataset sizes.

\subsection{Re-ranking For Reid}
Re-ranking methods, as a post-processing step in image retrieval, can significantly improve retrieval performance. To date, numerous effective re-ranking methods have been proposed. The k-reciprocal method first identifies the k most similar features within the test set: mutual nearest neighbors. It then merges these mutual nearest features and computes the Jaccard distance. The k-reciprocal method \cite{zhong2017reranking} typically consumes significant time; therefore, a novel GCN-based re-ranking method \cite{zhang2020understanding} was proposed. This method models the query and gallery as a whole and weights the K most similar image features from both the query and gallery, thereby reducing computation time. The Expanded Cross Neighborhood (ECN) distance was proposed \cite{sarfraz2018pose}, which does not require strict ranking list comparisons. ECN handles the query and gallery sets separately, calculating only the distance between each image and its neighboring images. This unsupervised method is more general and can be applied to image and video domains. A novel re-ranking method was proposed, which treats the ranking problem as a binary classification task \cite{zhou2022moving}. By combining local neighborhood information with the classifier's prediction results, the retrieval accuracy is improved. The AIM method \cite{fang2023semantic} was proposed, which not only leverages the correlation between the query and gallery to adjust the distance but also utilizes the relationships between gallery images to reduce noise within the gallery set, thereby refining the distance between the query and gallery.
\section{Methodology}
\label{sec:intro2}
In this section, we first introduce our proposed network architecture pipeline during both the training and testing phases. Then, we explain the HHCR optimization algorithm from both cross-modal and intra-modal perspectives, followed by a discussion of the similarity matrix after applying HHCR.

\subsection{Network Architecture Overview }
Formally, given a probe set visible images feature ${F_{v} =\{vi | i = 1, 2,... ..., N\}}$and a infrared set ir images feature with $N$ images ${F_{r} =\{ri | i = 1, 2,... ..., N\}}$, the cosine similarity between $v$ and $r$ as $F^{ori}_{sim} = \cos(F_{r},F_{v})$.The original similarity matrix $F^{ori}_{sim}$ and the overall similarity matrix $F_{sim}$ are as follows:
\begin{equation}
  F_{sim}=\cos({cat(F_{v},F_{r})},{cat(F_{v},F_{r})})
  \label{eq:also-important}
\end{equation}
\begin{equation}
F_{sim} = 
\begin{bmatrix}
{F}_{vv} , {F}_{vr} \\
{F}_{rv} ,{F}_{rr}
\end{bmatrix}
\end{equation}
Whereas $cos$ represents the cosine similarity, and $cat$ stands for matrix concatenation.${F}_{vv}$ denotes the similarity between visible images,${F}_{vr}$ denotes the similarity between visible and infrared images,${F}_{rv}$ denotes the similarity between infrared and visible images,${F}_{rr}$ denotes the similarity between infrared and infrared images.

Figure 3 illustrates the overall framework of our proposed baseline, which uses a single-stream ResNet network as the backbone. In the training phase, the network is constrained by the triplet loss function \cite{hermans2017defense} and the cross-entropy loss function \cite{luo2019bag}. During the testing phase, our proposed HHCR method is applied to explore the relationships of four matrix pairs:$\begin{bmatrix}{F}_{vv},{F}_{vr}\end{bmatrix}$,$\begin{bmatrix}{F}_{rv},{F}_{rr}\end{bmatrix}$, $\begin{bmatrix}{F}_{rr}\end{bmatrix}$, and $\begin{bmatrix}{F}_{vv}\end{bmatrix}$. The pairs $\begin{bmatrix}{F}_{vv},{F}_{vr}\end{bmatrix}$,$\begin{bmatrix}{F}_{rv},{F}_{rr}\end{bmatrix}$ are used to explore cross-modal heterogeneous relationships, while $\begin{bmatrix}{F}_{rr}\end{bmatrix}$, and $\begin{bmatrix}{F}_{vv}\end{bmatrix}$are used to explore intra-modal homogeneous relationships. HHCR jointly optimizes from both heterogeneous and homogeneous perspectives to find the most matched pairs in the visible and infrared sets.
\subsection{HHCR}
Our proposed HHCR adopts a progressive approach to optimize the ranking list. HHCR consists of two stages: heterogeneous re-ranking and homogeneous consistency re-ranking. In the first stage, HHCR searches for the k1/k4 most similar images from the query and gallery sets to the query/gallery. Then, it selects the k2/k5 most similar images from the query and gallery for query expansion. In the second stage, HHCR searches for the k2/k5 most similar images from the query/gallery and then finds the k3/k6 most similar images from the query for further query expansion. 

\textbf{Heterogeneous Consistency Re-ranking}

Due to the asymmetry between the query and gallery datasets, we split matrix M into two sub-matrices: ${F}_{v}^{sub}=\begin{bmatrix}{F}_{vv},{F}_{vr}\end{bmatrix}$ and ${F}_{r}^{sub}=\begin{bmatrix}{F}_{rv},{F}_{rr}\end{bmatrix}$, process ${F}_{v}^{sub}$ and ${F}_{r}^{sub}$ separately. Rank the top k1/k4 most similar elements for each query and gallery set from ${F}_{v}^{sub}/{F}_{r}^{sub}$.Define ${T}(k1,{F}_{v}^{sub})$  as the set of k1 elements with the highest ranking in ${F}_{v}$ set.Similarity, ${T}(k4,{F}_{r}^{sub})$  as the set of k4 elements with the highest ranking in ${F}_{r}^{sub}$ set.Assign different weights to the elements in $F_{sim}$ based on whether they belong to set ${T}(k1,{F}_{v}^{sub})$ or  ${T}(k4,{F}_{r}^{sub})$.

\begin{equation}
  F_{v}^{k1}=\begin{cases}1&f_i \in {T}(k1,{F}_{v}^{sub})\\0&f_i\notin{T}(k1,{F}_{v}^{sub})\end{cases}
  \label{eq:also-important2}
\end{equation}
\begin{equation}
  F_{r}^{k4}=\begin{cases}1&f_i \in{T}(k4,{F}_{r}^{sub})\\0&{f}_{i}\notin {T}(k4,{F}_{r}^{sub})\end{cases}
  \label{eq:also-important3}
\end{equation}
Where $f_i$ represents the elements in matrix $F_{sim}$,$A_{v}^{k1}$ and $ A_{r}^{k4}$ and represents the local adjacency matrix of query and gallery generated based on $F_{sim}$.

Combine matrices $A_{v}^{k1}$ and $ A_{r}^{k4}$ into matrix $W$, which is used as the adjacency matrix in the graph convolutional network. The adjacency matrix addresses the issue caused by the unequal number of visible and infrared images and assigns higher weight values to the most similar elements between the two sets.

\begin{equation}
  W=[F_{v}^{k1},F_{r}^{k4}]
\begin{tikzpicture}
    \draw (0,0) circle(0.1);
    \draw (0.0, 0.1) -- (0.0, -0.1);
    \draw (-0.1, 0.0) -- (0.1, 0.0);
\end{tikzpicture}
{[F_{v}^{k1},F_{r}^{k4}]}^T
  \label{eq:also-important4}
\end{equation}
 \noindent Where \begin{tikzpicture}
    \draw (0,0) circle(0.1);
    \draw (0.0, 0.1) -- (0.0, -0.1);
    \draw (0.1, 0.0) -- (-0.1, -0.0);
\end{tikzpicture}
denotes element-wise addition.

Rank all gallery elements in matrices ${M}_{v}/{M}_{r}$ for each query, denoted as ${F_{v}^{k1}}$ and ${F_{r}^{k4}}$. $F = F_{v}^{k1}+ F_{r}^{k4}$, which contains the candidate features of all visible and infrared images.Performing local queries on matrices ${F_{v}^{k1}}$ and ${F_{r}^{k4}}$ ensures that each visible and infrared image in matrix $F$ propagates information within its respective spatial neighborhood. This method helps mitigate the impact of abnormal features across modalities. We use matrix $W$ and the matrix $lqe(F)$ after information propagation as the adjacency matrix and node feature matrix in the graph convolutional network and aggregate them together. This operation helps explore the neighborhood relationships of each node within the node feature matrix.

\begin{equation}
  F_{v}^{k2}=\begin{cases}f_i&f_i \in {T}(k2,{M}_{v}^{sub})\\0&m_i\notin{T}(k2,{M}_{v}^{sub})\end{cases}
  \label{eq:also-important5}
\end{equation}
\begin{equation}
  F_{r}^{k5}=\begin{cases}f_i&f_i \in{T}(k5,{M}_{r}^{sub})\\0&{f}_{i}\notin {T}(k5,{M}_{r}^{sub})\end{cases}
  \label{eq:also-important6}
\end{equation}
\begin{equation}
  F^{lqe}_v,F^{lqe}_r = split(lqe(F)\begin{tikzpicture}
    \draw (0,0) circle(0.1);
    \draw (-0.07, 0.07) -- (0.07, -0.07);
    \draw (0.07, 0.07) -- (-0.07, -0.07);
\end{tikzpicture}W)
  \label{eq:also-important7}
\end{equation}
Whereas lqe represents the local query expansion operation, the split operation refers to dividing the aggregated matrix into visible and infrared parts, as $F_v^{lqe}$ and $F_r^{lqe}$, and cos represents using the cosine function to compute the similarity between the visible and infrared image sets.Specifically, $k1/k4 > k2/k5$.

\textbf{Homogeneous Consistency Reranking}

Homogeneous consistency re-ranking refers to minimizing the feature differences of the same pedestrian within the same modality. homogeneous consistency re-ranking is performed after the filtering by heterogeneous consistency re-ranking. We consider  $k1$ and $k4$ to represent the similar images from the visible and infrared modalities within the test set. Meanwhile, $k2$ and $k5$ represent the most similar images in their respective modality. Specifically, we believe that $k1$ and k4 include some images from the other modality, whereas $k2$ and $k5$ mostly consist of images from the same modality. Therefore, to reduce the impact of outlier images within the modality, Homogeneous Consistency Reranking further filters the effective information within the modality after selecting the top $k2$ and $k5$ similar images.

First, in matrices ${F}_{vv}$ and ${F}_{rr}$, the $k2/k5$ most similar elements are denoted as ${T}(k2,{F}_{v}^{sub}))$ and ${T}(k5,{F}_{r}^{sub}))$, the $k3/k6$ most similar elements are denoted as ${T}(k3,{F}_{v})$ and ${T}(k6,{F}_{r}^{sub}))$.
\begin{equation}
  A_{v}^{k2}=\begin{cases}1&f_i \in {T}(k2,{M}_{v}^{sub})\\0&f_i\notin{T}(k2,{M}_{v}^{sub}))\end{cases}
  \label{eq:also-important8}
\end{equation}
\begin{equation}
  A_{r}^{k5}=\begin{cases}1&f_i \in{T}(k5,{M}_{r}^{sub}))\\0&{f}_{i}\notin {T}(k5,{M}_{r}^{sub}))\end{cases}
  \label{eq:also-important9}
\end{equation}

\begin{equation}
  A_{v}^{k3}=\begin{cases}f_i&f_i \in {T}(k3,{M}_{v}^{sub}))\\0&f_i\notin{T}(k3,{M}_{v}^{sub}))\end{cases}
  \label{eq:also-important10}
\end{equation}
\begin{equation}
   A_{r}^{k6}=\begin{cases}f_i&f_i \in{T}(k6,{M}_{r}^{sub}))\\0&{f}_{i}\notin {T}(k6,{M}_{r}^{sub}))\end{cases}
  \label{eq:also-important11}
\end{equation}

Apply local query expansion to the obtained matrices, resulting in matrices $F_v^{filter}$ and $F_r^{filter}$. Matrices generated after lqe can filter out noise from homogeneous information within the same modality, bringing closer images of the same identity and pushing apart those of different identities.
\begin{equation}
  F_v^{filter}=lqe(F_{v}^{k3}\begin{tikzpicture}
    \draw (0,0) circle(0.1);
    \draw (-0.07, 0.07) -- (0.07, -0.07);
    \draw (0.07, 0.07) -- (-0.07, -0.07);
\end{tikzpicture}F_{v}^{k5})
  \label{eq:also-important12}
\end{equation}
\begin{equation}
  F_r^{filter}=lqe(F_{r}^{k3}\begin{tikzpicture}
    \draw (0,0) circle(0.1);
    \draw (-0.07, 0.07) -- (0.07, -0.07);
    \draw (0.07, 0.07) -- (-0.07, -0.07);
\end{tikzpicture}F_{r}^{k6})
  \label{eq:also-important1
  3}
\end{equation}
\begin{equation}
  \widetilde F_v^{rank}=F_v^{filter}\begin{tikzpicture}
    \draw (0,0) circle(0.1);
    \draw (-0.07, 0.07) -- (0.07, -0.07);
    \draw (0.07, 0.07) -- (-0.07, -0.07);
\end{tikzpicture}F_v^{rank}
  \label{eq:also-important14}
\end{equation}
\begin{equation}
  \widetilde F_r^{rank}=F_r^{filter}\begin{tikzpicture}
    \draw (0,0) circle(0.1);
    \draw (-0.07, 0.07) -- (0.07, -0.07);
    \draw (0.07, 0.07) -- (-0.07, -0.07);
\end{tikzpicture}F_r^{rank}
  \label{eq:also-important15}
\end{equation}
\begin{equation}
  \widetilde F_v=F_v^{filter}\begin{tikzpicture}
    \draw (0,0) circle(0.1);
    \draw (-0.07, 0.07) -- (0.07, -0.07);
    \draw (0.07, 0.07) -- (-0.07, -0.07);
\end{tikzpicture}F_v
  \label{eq:also-important16}
\end{equation}
\begin{equation}
  \widetilde F_r=F_r^{filter}\begin{tikzpicture}
    \draw (0,0) circle(0.1);
    \draw (-0.07, 0.07) -- (0.07, -0.07);
    \draw (0.07, 0.07) -- (-0.07, -0.07);
\end{tikzpicture}F_r
  \label{eq:also-important17}
\end{equation}
where as $F_v^{filter}$ and $F_r^{filter}$ represent the filtered results generated after the LQE operation,$ \widetilde F_v^{rank}$ a represents the matrix $F_v^{rank}$ after the removal of noise within the modality,$\widetilde F_r^{rank}$ a represents the matrix $ F_r^{rank}$ after the removal of noise within the modality,$\widetilde F_r$ a represents the matrix $F_r$ after the removal of noise within the modality,$\widetilde F_v$ a represents the matrix $F_v$ after the removal of noise within the modality.

\textbf{Final Similarity Matrix}

The final similarity matrix comprises a weighted combination of the original similarity matrix and the matrix obtained after re-ranking. Fixed parameter $\lambda$ is used to control the weights of the two matrices. Heterogeneous consistency is responsible for exploring the relationships between the test sets from a global perspective. homogeneous consistency re-ranking focuses on eliminating noise within the query and gallery sets.

The final similarity matrix can be represented by Equation (16).
\begin{equation}
\hat{F}^{final}_{sim}= (1-\lambda)\widetilde F_v^{rank}*\widetilde F_r^{rank}+\lambda \widetilde  F_v* \widetilde F_r
\end{equation}

\section{Experiments}
\label{sec:intro3}

\subsection{Dataset}

\textbf{LLCM:} This dataset \cite{diverse_embedding_expansion_network} was collected using 18 cameras, including nine visible-light cameras and nine infrared cameras. It is currently the largest infrared-visible dataset, containing a total of 46,767 images of 1,064 individuals. This dataset is particularly challenging due to the inclusion of various illumination conditions, which increases the difficulty of pedestrian recognition.

\textbf{SYSU-MM01:} This large-scale dataset \cite{wu2017rgb} was collected at Sun Yat-sen University, using six cameras in total—four visible-light cameras and two infrared cameras, covering both indoor and outdoor environments. The dataset contains 303,420 images of 491 individuals, with each identity represented in both modalities. During testing, the dataset can be used in two modes: indoor search and all-search.

\textbf{RegDB:} This dataset \cite{nguyen2017person} was collected using dual cameras and contains 8,240 images of 412 individuals, with each identity having ten infrared and ten visible-light images. In this dataset, each infrared image is paired with a corresponding visible-light image, and it has less variation in pedestrian poses compared to SYSU-MM01, making it easier to detect.

\subsection{ Implementation Details}

Following saai, we adopt a ResNet network pre-trained on ImageNet as the backbone. For each batch, we select 10 identities, with 8 images per identity. Each image is resized to 288 × 144. The network is optimized using the Adam optimizer with a linear warmup strategy. The initial learning rate is set to 3.5 × 10-4 and is reduced by factors of 0.1 and 0.01 at 80 and 120 epochs, respectively. The training process spans a total of 160 epochs. 
\subsection{Performance of the HHCR Method }

In this section, we evaluate the HHCR method's superiority and generalizability. First, we assess the CMBR method's matching performance compared to other reranking methods on the same dataset and model. Then, we test the generalizability of our proposed method by applying it to other models in the field of person re-identification. 

\textbf{Superiority Of The HHCR Method:} In the experiments evaluating superiority, we maintained the same settings as GNN. The experiments were conducted on the RegDB, SYSU, and LLCM datasets. As shown in Table 1, we compared our proposed method with other re-ranking methods on cross-modality person re-identification datasets. In these experiments, we primarily observed two metrics: Rank-1 accuracy and mAP.The loss function consists of cross-entropy loss and triplet loss. Extensive experiments have demonstrated that the proposed PMRR method has achieved the SOTA. In the SYSU dataset, Rank-1 reached 83.98\%, and mAP reached 85.32\%. In the RegDB dataset, PMRR achieved the highest accuracy, with Rank-1 reaching 89.3\% and mAP reaching 89.9\%. In the LLCM dataset, Rank-1 reached 80.6\%, and mAP reached 76.8\%.

\begin{table}[H]
  \centering
  \scriptsize
  \renewcommand{\arraystretch}{1.5} 
  \caption{Comparison of Other Re-Ranking Methods and HHCR on Various Cross-Modal Person Re-Identification Datasets.}
  \begin{tabular}{@{}l|c@{\hspace{5pt}}c|c@{\hspace{5pt}}c|c@{\hspace{5pt}}c}
    \hline
    \multirow{2}{*}{\centering methods} &\multicolumn{2}{c|}{SYSU} &\multicolumn{2}{c|}{RegDB}& \multicolumn{2}{c}{LLCM}\\ 
    \cline{2-7}
           & Rank1& mAP1& Rank1& mAP1& Rank1& mAP1  \\
    \hline
    k-reciprocal      & 83.22  & 80.60 &86.75 & 87.41       & 72.70   & 70.99      \\ 
    GNN  &81.96&83.36   &86.31&89.30      & 76.19    & 75.43     \\ 
    ECN     &83.93&85.04&    87.28   &   88.63   & 74.88    & 72.89       \\ 
    Ours   &\textbf{83.98} &\textbf{85.32}&\textbf{89.3} &\textbf{89.9}&   \textbf{80.6}&  \textbf{76.8}      \\ 
    \hline
  \end{tabular}

  \label{tab:example15}
\end{table}
\begin{table}[H]
  \centering
  \scriptsize
  \renewcommand{\arraystretch}{1.5} 
  \caption{Comparison of Applying HHCR and Without HHCR in Other V-I ReID Models.}
  \begin{tabular}{@{}l|c@{\hspace{5pt}}c|c@{\hspace{5pt}}c|c@{\hspace{5pt}}c}
    \hline
    \multirow{2}{*}{\centering methods}  & \multicolumn{2}{c|}{SYSU} & \multicolumn{2}{c|}{RegDB} & \multicolumn{2}{c}{LLCM}\\
    \cline{2-7}
            & mAP(\%) & Rank1(\%) & mAP(\%) & Rank1(\%) & mAP(\%) & Rank1(\%)   \\
    \hline
    AGW  & 87.75   &  88.44         &    88.59     &     90.50     &      -              &    -  \\
    Ours   &87.31  &     88.26    &    96.54   &     97.19     &             -        &    -    \\
    Yours  & -    &      -     &     -     &      -     &         -            &    -\\
    Ours   & -    &       -    & -         &      -     &          -           &       -\\
    \hline
  \end{tabular}

  \label{tab:example27}
\end{table}

\begin{table*}[ht]
  \centering
  \scriptsize
  \renewcommand{\arraystretch}{1.5} 
  \caption{Comparison with State-of-the-Art Methods on SYSU-MM01.}
  \begin{tabular}{@{}l|cc|cc|cc|cc}
        \hline
    \multirow{3}{*}{\centering Methods}& \multicolumn{4}{c|}{All-Search}& \multicolumn{4}{c}{Indoor-Search} \\ 
    \cline{2-9}
 					& \multicolumn{2}{c|}{Single-Shot} & \multicolumn{2}{c|}{Multi-Shot}& \multicolumn{2}{c|}{Single-Shot} & \multicolumn{2}{c}{Multi-Shot} \\ 
    \cline{2-9}
					 & Rank1     & mAP & Rank1     & mAP& Rank1     & mAP  & Rank1     & mAP   \\ 
    \hline
   
    Zero-Padding\cite{wu2017rgb}          &14.80& 15.95 &19.13& 10.89 &20.58 &26.92& 24.43& 18.86 \\
    cmGAN\cite{dai2018cross}                    &26.97 &27.80 &31.49 &22.27& 31.63 &42.19& 37.00 &32.76\\
    JSIA-ReID\cite{wang2020paired}    		&38.10& 36.90& 45.10 &29.50 &43.80& 52.90&52.70 &42.70\\
    AlignGAN\cite{wang2019joint}  		&42.40 &40.70& 51.50& 33.90& 45.90 &54.30& 57.10& 45.30\\
    AGW\cite{ye2021deep}   			&47.50& 47.65& -& -& 54.1&62.97& -& -\\
    LbA\cite{park2021learning}  			&55.41 &54.14& -& - &58.46 &66.33 &-& - \\
    NFS\cite{chen2021neural}   			&56.91& 55.45& 63.51& 48.56& 62.79& 69.79& 70.03 &61.45\\
    MID\cite{huang2022modality} 			& 60.27 &59.40 &-& -& 64.86& 70.12& -& -\\
    cm-SSFT\cite{lu2020cross} 		& 61.60& 63.20& 63.40& 62.00& 70.50& 72.60 &73.00& 72.40\\
    CM-NAS\cite{fu2021cmnas}  		& 61.99 &60.02 &68.68& 53.45 &67.01 &72.95& 76.48 &65.11\\
    MCLNet\cite{nguyen2017person}   		&65.40& 61.98 &-& -& 72.56& 76.58& -& - \\
    FMCNet\cite{zhang2022fmcnet} 		& 66.34& 62.51 &73.44 &56.06 &68.15& 74.09 &78.86 &63.82 \\
    SMCL\cite{wei2021syncretic}   			&67.39 &61.78& 72.15& 54.93 &68.84 &75.56& 79.57& 66.57\\
    MPANet\cite{vaswani2017attention}  		&70.58 &68.24& 75.58 &62.91& 76.74 &80.95& 84.22& 75.11\\
    MAUM\cite{liu2022memory} 			& 71.68 &68.79 &-& - &76.97& 81.94 &-& -\\
    CMT\cite{jiang2022transformer}  			&71.88& 68.57& 80.23& 63.13& 76.90& 79.91 &84.87 &74.11\\
    CIFT\cite{li2022counterfactual}  			&74.08 &74.79& 79.74 &75.56& 81.82& 85.61& 88.32& 86.42\\
    MSCLNet\cite{zhang2022modality}  		&76.99 &71.64 &- &- &78.49& 81.17& - &-\\
    SAAI\cite{fang2023semantic}  			&75.90& 77.03& 82.86 &82.39& 83.20& 88.01& 90.73& 91.30\\
    ours  			&\textbf{83.99}& \textbf{85.32}&\textbf{87.44} &\textbf{87.63}& \textbf{90.42}&\textbf{93.31}&\textbf{94.76}&\textbf{95.16}\\
    \bottomrule
  \end{tabular}

  \label{tab:example3}
\end{table*}

\textbf{Retrieval Performance on other V-I ReID Net:}
After extracting features using current state-of-the-art cross-modality person re-identification models, we apply our proposed reranking method to match these features. Specifically, we replace the Euclidean distance and similarity matrix used in other models' feature matching with our CMBR algorithm. Despite this,  As shown in Table 2, the accuracy of numerous models saw significant improvement after incorporating our reranking algorithm.

\subsection{Comparison With State-of-the-art Methods}
We compare our baseline with current state-of-the-art methods to demonstrate the superiority of our approach. Experiments on the SYSU are presented in Table 3, experiments on the RegDB are presented in Table 4, and experiments on the LLCM dataset are shown in Table 5.

\textbf{Comparison on SYSU-MM01}As shown in Table 3, the proposed baseline has achieved state-of-the-art performance. Our model achieved a Rank-1 accuracy of 76.6\% and a mAP of 82.0\% in the all-search Single-Shot mode. In the all-search Multi-Shot mode, our model achieved a Rank-1 accuracy of 88.9\% and a mAP of 89.3\%. In the indoor-search Single-Shot mode, our model achieved a Rank-1 accuracy of 88.0\% and a mAP of 91.5\%. Finally, in the indoor-search Multi-Shot mode, our model achieved a Rank-1 accuracy of 94.4\% and a mAP of 95.0\%.

\textbf{Comparison on RegDB}
As shown in Table 4, the proposed baseline has achieved state-of-the-art performance. In the Visible to Infrared mode, our model achieved a Rank-1 accuracy of ****\% and an mAP of ***\%. In the Infrared to Visible mode, our model reached a Rank-1 accuracy of ****\% and an mAP of ****\%.

\textbf{Comparison on LLCM}
As shown in Table 5, the proposed baseline has achieved state-of-the-art performance. In the Visible to Infrared mode, our model achieved a Rank-1 accuracy of ****\% and an mAP of ***\%. In the Infrared to Visible mode, our model reached a Rank-1 accuracy of ****\% and an mAP of ****\%.

\begin{table}[ht]
  \centering
  \scriptsize
  \renewcommand{\arraystretch}{1.5} 
  \caption{Comparison with State-of-the-Art Methods on Regdb.}
  \begin{tabular}{l|cc|cc}
    \hline
    \multirow{2}{*}{\centering methods}  & \multicolumn{2}{c|}{VIS to IR} & \multicolumn{2}{c}{IR to VIS}\\
    \cline{2-5}
            & Rank1 & mAP& Rank1 & mAP1 \\
    \hline
    Zero-Padding\cite{wu2017rgb} & 17.75 &18.90&16.63 &17.82      \\
    JSIA-ReID\cite{wang2020paired} & 48.50& 49.30& 48.10& 48.90\\
   AlignGAN\cite{wang2019joint} & 57.90 &53.60 &56.30& 53.40\\
    AGW\cite{ye2021deep} & 70.05 &66.37& 70.49& 65.9 \\
    cm-SSFT\cite{lu2020cross} & 72.30& 72.90 &71.00 &71.70     \\
    LbA\cite{park2021learning}  &74.17& 67.64 &72.43& 65.46   \\
    MCLNet\cite{nguyen2017person}& 80.31& 73.07 &75.93& 69.4     \\
    NFS\cite{chen2021neural} & 80.54 &72.10 &77.95& 69.79   \\
   MPANet\cite{vaswani2017attention} &83.70& 80.90 &82.80& 80.70 \\
    SMCL\cite{wei2021syncretic}& 83.93 &79.83& 83.05 &78.57     \\
    MSCLNet\cite{zhang2022modality}&84.17& 80.99 &83.86& 78.31   \\
   CM-NAS\cite{fu2021cmnas}  &84.54 &80.32& 82.57& 78.31     \\
    MID\cite{huang2022modality}  &87.45 &84.85 &84.29& 81.41     \\
    MAUM\cite{liu2022memory}  &87.87& 85.09& 86.95& 84.3     \\
    FMCNet\cite{zhang2022fmcnet} &89.12& 84.43& 88.38& 83.86\\
    CIFT\cite{li2022counterfactual} &91.96 &92.00& 90.30& 90.7    \\
    CMT\cite{jiang2022transformer}  &95.17 &87.30 &91.97& 84.46    \\
    SAAI\cite{fang2023semantic} &91.07 &91.45& 92.09 &92.0\\
    ours &\textbf{90.63} &\textbf{92.83}& \textbf{92.52}& \textbf{94.26}\\
    \hline
  \end{tabular}
  \label{tab:example42}
\end{table}
\begin{table}[ht]
  \centering
  \scriptsize
  \renewcommand{\arraystretch}{1.5} 
  \caption{Comparison with State-of-the-Art Methods on LLCM.}
  \begin{tabular}{@{}l|cc|cc@{}}
    \hline
    \multirow{2}{*}{\centering methods} & \multicolumn{2}{c|}{IR to VIS} & \multicolumn{2}{c}{VIS to IR} \\
    \cline{2-5}
            & Rank1& mAP& Rank1& mAP\\
    \hline
    DDAG \cite{wang2019joint}& 40.3 & 48.4 &48.0  &52.3\\
    AGW \cite{ye2021deep} &43.6 &51.8 &51.5 & 55.3 \\
    LbA \cite{park2021learning}& 43.8 &53.1 &50.8&55.6 \\
    CAJ \cite{ye2021channel} &48.8 &56.6& 56.5& 59.8\\
    DART \cite{yang2022twin}& 52.2& 59.8& 60.4 &63.2 \\
    MMN \cite{zhang2021unified} &52.5 &58.9& 59.9& 62.7\\
    DEEN\cite{diverse_embedding_expansion_network} & 54.9& 62.9& 62.5& 65.8\\
    ours & \textbf{75.87}&\textbf{75.24}& \textbf{82.33} & \textbf{80.00}\\
    \hline
  \end{tabular}

  \label{tab:example5}
\end{table}

\subsection{Ablation Study}

We use the model without any reranking algorithms added in this paper as the baseline. In the experiments, we fixed lambda at 0.3 and separately tested the impact of Intrinsic Consistency Reranking and Hetero-spectral Reranking on the baseline. We also tested the effects of the adjacency matrix Transposition Filtering and symmetric critical matrix Ranked Transposition Filtering on the model. As shown in Table 6, The experiments show that our proposed reranking algorithm is effective and can significantly improve the model's accuracy. It is particularly noteworthy that HR w/o RTF achieved the highest performance on the RegDB and LLCM datasets but only reached Rank-1 accuracy of 80.49\% and mAP of 82.59\% on the SYSU dataset. On the other hand, HR RTF achieved the highest accuracy on the SYSU dataset but did not attain the best performance on the RegDB and LLCM datasets. When comparing with other re-ranking methods, it was found that HR w/o RTF combined with HHCR showed relatively poor performance on SYSU. However, HR RTF consistently outperformed other re-ranking methods across all three datasets. Therefore, the final HHCR is based on the HR RTF architecture rather than HR w/o RTF.

\begin{table}[ht]
  \centering
  \scriptsize
  \renewcommand{\arraystretch}{1.5} 
  \caption{The impact of Ranked Transposition Filtering, Transposition Filtering, Intrinsic Consistency Reranking, and Hetero-spectral Re-ranking on performance.}
  \begin{tabular}{l@{\hspace{5pt}}c@{\hspace{5pt}}c|c@{\hspace{5pt}}c|c@{\hspace{5pt}}c|c@{\hspace{5pt}}c}
    \hline

    \multirow{2}{*}{\centering ICR}&\multirow{2}{*}{\centering HR}&\multirow{2}{*}{\centering HR w/o RTF}& \multicolumn{2}{c|}{RegDB} & \multicolumn{2}{c|}{SYSU} & \multicolumn{2}{c}{LLCM}\\
	\cline{4-9}
        &&  &  Rank1 & mAP& Rank1 & mAP1&Rank1 & mAP1\\
    \hline
     &&& 86.3&83.7&66.18&64.31&63.09&49.07\\
     \checkmark&&&85.6 &89.0&64.34&72.02&71.1&66.57  \\

     \checkmark&\checkmark&& 90.7 &92.9&\textbf{83.98}&\textbf{85.32}&75.87&75.24  \\
     \checkmark&&\checkmark& \textbf{91.8} &\textbf{93}&80.49&82.59&\textbf{80.6}&\textbf{76.8}  \\
        \hline
  \end{tabular}

  \label{tab:example41}
\end{table}

\subsection{HHCR result Visualization}
 \begin{figure}[ht]
  \centering
   \includegraphics[width=1\linewidth]{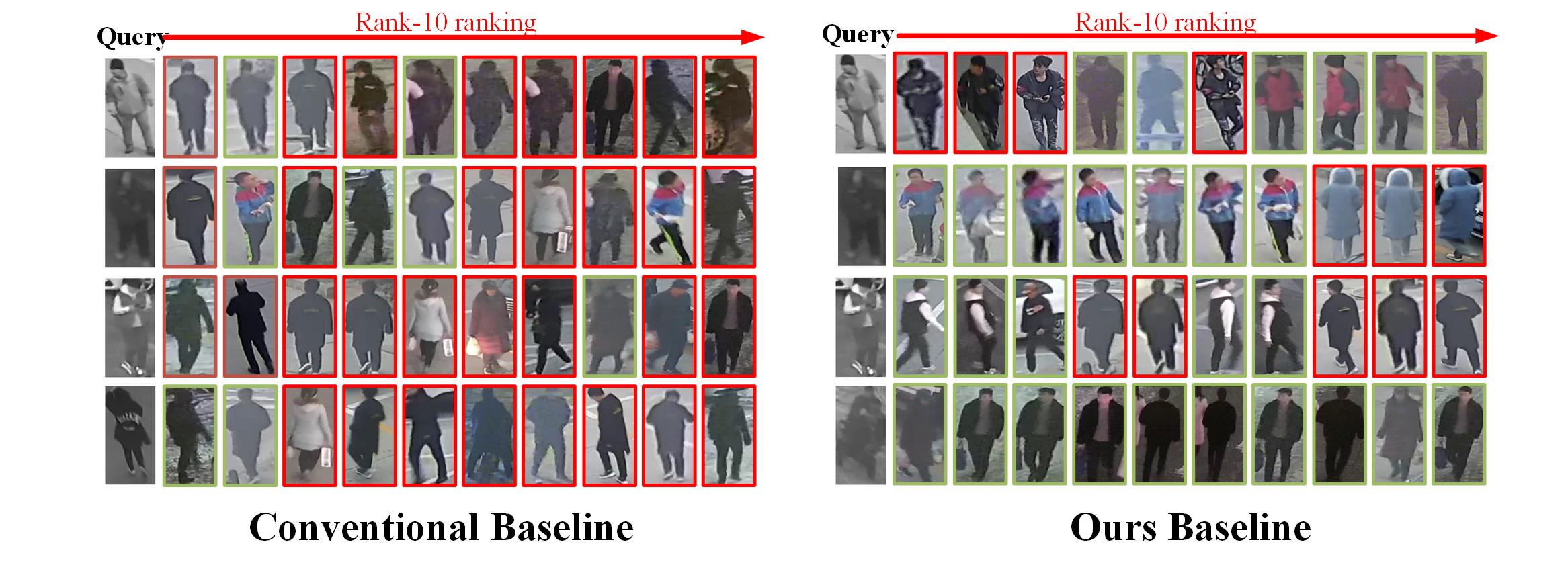}

   \caption{Comparison of Top 10 Retrieval Results with and without PMRR.}
   \label{fig:onecol122}
\end{figure}
To make the retrieval results of our HHCR method more clearly visible, we compared the visualized results with and without the application of HHCR. In Figure 3, the image on the far left represents the query image, while the images on the right represent the top ten most similar retrieval results. Green indicates identity matches, and red indicates mismatches. The experimental results show that, in the top ten ranks, the number of correctly retrieved images significantly increases after applying HHCR compared to the results without HHCR.

\section{Conclusion}
First, we propose a novel baseline for Cross-modal Person Re-identification. Then, we introduce a superior and generalizable HHCR feature--the matching method, which is a two-stage reranking approach. The first stage is Hetero-spectral  Reranking, which reduces the differences between modalities. The second stage is Intrinsic Consistency Reranking, which minimizes the differences within each modality. This two-stage reranking process significantly improves the accuracy of multi--modal feature matching. Extensive experiments on the SYSU-MM01, LLCM, and RegDB datasets demonstrate the effectiveness of our method.

\bibliographystyle{IEEEtran}
\bibliography{reference}

\begin{thebibliography}{10}
\providecommand{\url}[1]{#1}
\csname url@samestyle\endcsname
\providecommand{\newblock}{\relax}
\providecommand{\bibinfo}[2]{#2}
\providecommand{\BIBentrySTDinterwordspacing}{\spaceskip=0pt\relax}
\providecommand{\BIBentryALTinterwordstretchfactor}{4}
\providecommand{\BIBentryALTinterwordspacing}{\spaceskip=\fontdimen2\font plus
\BIBentryALTinterwordstretchfactor\fontdimen3\font minus
  \fontdimen4\font\relax}
\providecommand{\BIBforeignlanguage}[2]{{%
\expandafter\ifx\csname l@#1\endcsname\relax
\typeout{** WARNING: IEEEtran.bst: No hyphenation pattern has been}%
\typeout{** loaded for the language `#1'. Using the pattern for}%
\typeout{** the default language instead.}%
\else
\language=\csname l@#1\endcsname
\fi
#2}}
\providecommand{\BIBdecl}{\relax}
\BIBdecl

\bibitem{quan2019autoreid}
R.~Quan, X.~Dong, Y.~Wu \emph{et~al.}, ``Auto-reid: Searching for a part-aware
  convnet for person re-identification,'' in \emph{Proceedings of the IEEE/CVF
  International Conference on Computer Vision (ICCV)}, 2019, pp. 3750--3759.

\bibitem{luo2019bag}
H.~Luo, Y.~Gu, X.~Liao \emph{et~al.}, ``Bag of tricks and a strong baseline for
  deep person re-identification,'' in \emph{Proceedings of the IEEE/CVF
  Conference on Computer Vision and Pattern Recognition Workshops}, 2019, pp.
  0--0.

\bibitem{huang2019celebrities}
Y.~Huang, Q.~Wu, J.~Xu \emph{et~al.}, ``Celebrities-reid: A benchmark for
  clothes variation in long-term person re-identification,'' in \emph{2019
  International Joint Conference on Neural Networks (IJCNN)}.\hskip 1em plus
  0.5em minus 0.4em\relax IEEE, 2019, pp. 1--8.

\bibitem{cui2023dcr}
Z.~Cui, J.~Zhou, Y.~Peng \emph{et~al.}, ``Dcr-reid: Deep component
  reconstruction for cloth-changing person re-identification,'' \emph{IEEE
  Transactions on Circuits and Systems for Video Technology}, vol.~33, no.~8,
  pp. 4415--4428, 2023.

\bibitem{wang2020paired}
G.~Wang, T.~Zhang, Y.~Yang \emph{et~al.}, ``Cross-modality paired-images
  generation for rgb-infrared person re-identification,'' in \emph{Proceedings
  of the AAAI Conference on Artificial Intelligence}, vol.~34, no.~7, 2020, pp.
  12\,144--12\,151.

\bibitem{hu2020maximum}
X.~Hu and Y.~Zhou, ``Cross-modality person reid with maximum intra-class
  triplet loss,'' in \emph{Pattern Recognition and Computer Vision: Third
  Chinese Conference, PRCV 2020}.\hskip 1em plus 0.5em minus 0.4em\relax
  Springer International Publishing, 2020, pp. 557--568.

\bibitem{vaswani2017attention}
A.~Vaswani \emph{et~al.}, ``Attention is all you need,'' \emph{Advances in
  Neural Information Processing Systems}, 2017.

\bibitem{wang2019joint}
G.~Wang, T.~Zhang, J.~Cheng \emph{et~al.}, ``Rgb-infrared cross-modality person
  re-identification via joint pixel and feature alignment,'' in
  \emph{Proceedings of the IEEE/CVF International Conference on Computer Vision
  (ICCV)}, 2019, pp. 3623--3632.

\bibitem{diverse_embedding_expansion_network}
``Diverse embedding expansion network and low-light cross-modality benchmark
  for visible-infrared person re-identification,'' incomplete bibliographic
  information; please add authors, venue, and year.

\bibitem{li2022counterfactual}
X.~Li, Y.~Lu, B.~Liu \emph{et~al.}, ``Counterfactual intervention feature
  transfer for visible-infrared person re-identification,'' in \emph{European
  Conference on Computer Vision (ECCV)}.\hskip 1em plus 0.5em minus 0.4em\relax
  Springer Nature Switzerland, 2022, pp. 381--398.

\bibitem{feng2023shape}
J.~Feng, A.~Wu, and W.-S. Zheng, ``Shape-erased feature learning for
  visible-infrared person re-identification,'' in \emph{Proceedings of the
  IEEE/CVF Conference on Computer Vision and Pattern Recognition (CVPR)}, 2023,
  pp. 22\,752--22\,761.

\bibitem{sarfraz2018pose}
M.~S. Sarfraz, A.~Schumann, A.~Eberle \emph{et~al.}, ``A pose-sensitive
  embedding for person re-identification with expanded cross neighborhood
  re-ranking,'' in \emph{Proceedings of the IEEE Conference on Computer Vision
  and Pattern Recognition (CVPR)}, 2018, pp. 420--429.

\bibitem{zhou2022moving}
Y.~Zhou, Y.~Wang, and L.-P. Chau, ``Moving towards centers: Re-ranking with
  attention and memory for re-identification,'' \emph{IEEE Transactions on
  Multimedia}, vol.~25, pp. 3456--3468, 2022.

\bibitem{fang2023semantic}
X.~Fang, Y.~Yang, and Y.~Fu, ``Visible-infrared person re-identification via
  semantic alignment and affinity inference,'' in \emph{Proceedings of the
  IEEE/CVF International Conference on Computer Vision (ICCV)}, 2023, pp.
  11\,270--11\,279.

\bibitem{wu2017rgb}
A.~Wu, W.-S. Zheng, H.-X. Yu \emph{et~al.}, ``Rgb-infrared cross-modality
  person re-identification,'' in \emph{Proceedings of the IEEE International
  Conference on Computer Vision (ICCV)}, 2017, pp. 5380--5389.

\bibitem{goodfellow2014gan}
I.~Goodfellow, J.~Pouget-Abadie, M.~Mirza \emph{et~al.}, ``Generative
  adversarial nets,'' \emph{Advances in Neural Information Processing Systems},
  vol.~27, 2014.

\bibitem{dai2018cross}
P.~Dai, R.~Ji, H.~Wang \emph{et~al.}, ``Cross-modality person re-identification
  with generative adversarial training,'' in \emph{IJCAI}, vol.~1, no.~3, 2018,
  p.~6.

\bibitem{ye2018visible}
M.~Ye, Z.~Wang, X.~Lan \emph{et~al.}, ``Visible thermal person
  re-identification via dual-constrained top-ranking,'' in \emph{IJCAI},
  vol.~1, 2018, p.~2.

\bibitem{ye2020dynamic}
M.~Ye, J.~Shen, D.~J. Crandall \emph{et~al.}, ``Dynamic dual-attentive
  aggregation learning for visible-infrared person re-identification,'' in
  \emph{Computer Vision -- ECCV 2020}.\hskip 1em plus 0.5em minus 0.4em\relax
  Springer International Publishing, 2020, pp. 229--247.

\bibitem{lu2020cross}
Y.~Lu, Y.~Wu, B.~Liu \emph{et~al.}, ``Cross-modality person re-identification
  with shared-specific feature transfer,'' in \emph{Proceedings of the IEEE/CVF
  Conference on Computer Vision and Pattern Recognition (CVPR)}, 2020, pp.
  13\,379--13\,389.

\bibitem{ye2021deep}
M.~Ye, J.~Shen, G.~Lin \emph{et~al.}, ``Deep learning for person
  re-identification: A survey and outlook,'' \emph{IEEE Transactions on Pattern
  Analysis and Machine Intelligence}, vol.~44, no.~6, pp. 2872--2893, 2021.

\bibitem{wang2018nonlocal}
X.~Wang, R.~Girshick, A.~Gupta \emph{et~al.}, ``Non-local neural networks,'' in
  \emph{Proceedings of the IEEE Conference on Computer Vision and Pattern
  Recognition (CVPR)}, 2018, pp. 7794--7803.

\bibitem{he2016deep}
K.~He, X.~Zhang, S.~Ren \emph{et~al.}, ``Deep residual learning for image
  recognition,'' in \emph{Proceedings of the IEEE Conference on Computer Vision
  and Pattern Recognition (CVPR)}, 2016, pp. 770--778.

\bibitem{zhong2017reranking}
Z.~Zhong, L.~Zheng, D.~Cao \emph{et~al.}, ``Re-ranking person re-identification
  with k-reciprocal encoding,'' in \emph{Proceedings of the IEEE Conference on
  Computer Vision and Pattern Recognition (CVPR)}, 2017, pp. 1318--1327.

\bibitem{zhang2020understanding}
X.~Zhang, M.~Jiang, Z.~Zheng \emph{et~al.}, ``Understanding image retrieval
  re-ranking: A graph neural network perspective,'' \emph{arXiv preprint
  arXiv:2012.07620}, 2020.

\bibitem{hermans2017defense}
A.~Hermans, L.~Beyer, and B.~Leibe, ``In defense of the triplet loss for person
  re-identification,'' \emph{arXiv preprint arXiv:1703.07737}, 2017.

\bibitem{nguyen2017person}
D.~T. Nguyen, H.~G. Hong, K.~R. Kim \emph{et~al.}, ``Person recognition system
  based on a combination of body images from visible light and thermal
  cameras,'' \emph{Sensors}, vol.~17, no.~3, p. 605, 2017.

\bibitem{park2021learning}
H.~Park, S.~Lee, J.~Lee \emph{et~al.}, ``Learning by aligning: Visible-infrared
  person re-identification using cross-modal correspondences,'' in
  \emph{Proceedings of the IEEE/CVF International Conference on Computer Vision
  (ICCV)}, 2021, pp. 12\,046--12\,055.

\bibitem{chen2021neural}
Y.~Chen, L.~Wan, Z.~Li \emph{et~al.}, ``Neural feature search for rgb-infrared
  person re-identification,'' in \emph{Proceedings of the IEEE/CVF Conference
  on Computer Vision and Pattern Recognition (CVPR)}, 2021, pp. 587--597.

\bibitem{huang2022modality}
Z.~Huang, J.~Liu, L.~Li \emph{et~al.}, ``Modality-adaptive mixup and invariant
  decomposition for rgb-infrared person re-identification,'' in
  \emph{Proceedings of the AAAI Conference on Artificial Intelligence},
  vol.~36, no.~1, 2022, pp. 1034--1042.

\bibitem{fu2021cmnas}
C.~Fu, Y.~Hu, X.~Wu \emph{et~al.}, ``Cm-nas: Cross-modality neural architecture
  search for visible-infrared person re-identification,'' in \emph{Proceedings
  of the IEEE/CVF International Conference on Computer Vision (ICCV)}, 2021,
  pp. 11\,823--11\,832.

\bibitem{zhang2022fmcnet}
Q.~Zhang, C.~Lai, J.~Liu \emph{et~al.}, ``Fmcnet: Feature-level modality
  compensation for visible-infrared person re-identification,'' in
  \emph{Proceedings of the IEEE/CVF Conference on Computer Vision and Pattern
  Recognition (CVPR)}, 2022, pp. 7349--7358.

\bibitem{wei2021syncretic}
Z.~Wei, X.~Yang, N.~Wang \emph{et~al.}, ``Syncretic modality collaborative
  learning for visible infrared person re-identification,'' in
  \emph{Proceedings of the IEEE/CVF International Conference on Computer Vision
  (ICCV)}, 2021, pp. 225--234.

\bibitem{liu2022memory}
J.~Liu, Y.~Sun, F.~Zhu \emph{et~al.}, ``Learning memory-augmented
  unidirectional metrics for cross-modality person re-identification,'' in
  \emph{Proceedings of the IEEE/CVF Conference on Computer Vision and Pattern
  Recognition (CVPR)}, 2022, pp. 19\,366--19\,375.

\bibitem{jiang2022transformer}
K.~Jiang, T.~Zhang, X.~Liu \emph{et~al.}, ``Cross-modality transformer for
  visible-infrared person re-identification,'' in \emph{European Conference on
  Computer Vision (ECCV)}.\hskip 1em plus 0.5em minus 0.4em\relax Springer
  Nature Switzerland, 2022, pp. 480--496.

\bibitem{zhang2022modality}
Y.~Zhang, S.~Zhao, Y.~Kang \emph{et~al.}, ``Modality synergy complement
  learning with cascaded aggregation for visible-infrared person
  re-identification,'' in \emph{European Conference on Computer Vision
  (ECCV)}.\hskip 1em plus 0.5em minus 0.4em\relax Springer Nature Switzerland,
  2022, pp. 462--479.

\bibitem{ye2021channel}
M.~Ye, W.~Ruan, B.~Du \emph{et~al.}, ``Channel augmented joint learning for
  visible-infrared recognition,'' in \emph{Proceedings of the IEEE/CVF
  International Conference on Computer Vision (ICCV)}, 2021, pp.
  13\,567--13\,576.

\bibitem{yang2022twin}
M.~Yang, Z.~Huang, P.~Hu \emph{et~al.}, ``Learning with twin noisy labels for
  visible-infrared person re-identification,'' in \emph{Proceedings of the
  IEEE/CVF Conference on Computer Vision and Pattern Recognition (CVPR)}, 2022,
  pp. 14\,308--14\,317.

\bibitem{zhang2021unified}
Y.~Zhang, Y.~Yan, Y.~Lu \emph{et~al.}, ``Towards a unified middle modality
  learning for visible-infrared person re-identification,'' in
  \emph{Proceedings of the 29th ACM International Conference on Multimedia},
  2021, pp. 788--796.

\end{thebibliography}

\end{document}